\documentclass{article}
\pdfoutput=1

\PassOptionsToPackage{numbers}{natbib}
\usepackage[preprint]{neurips_2023}

\usepackage[utf8]{inputenc} % allow utf-8 input
\usepackage[T1]{fontenc}    % use 8-bit T1 fonts
\usepackage{hyperref}       % hyperlinks
\usepackage{url}            % simple URL typesetting
\usepackage{booktabs}       % professional-quality tables
\usepackage{amsfonts}       % blackboard math symbols
\usepackage{nicefrac}       % compact symbols for 1/2, etc.
\usepackage{graphicx}       % figures
\usepackage{amsmath}        % math-features
\usepackage{adjustbox}      % adjust table scale
\usepackage{microtype}      % microtypography
\usepackage{xcolor}         % colors
\usepackage{caption}        % figures
\usepackage{subcaption}     % and subfigures
\usepackage{multirow}       % tables
\usepackage{xcolor}
\usepackage{listings}
\usepackage{cleveref}       % 
\title{Inductive Bias for Emergent Communication in a Continuous Setting}

\author{%
  John Isak Fjellvang Villanger  \\
  Department of Energy and Petroleum Engineering,\\
  University of Stavanger, Norway\\
  \texttt{john.i.villanger@uis.com} \\
  \And
  Troels Arnfred Bojesen \\
  Department of Informatics,\\
  University of Bergen, Norway\\
  \texttt{Troels.Bojesen@uib.no} \\
}

\begin{document}

\maketitle

\begin{abstract}
    We study emergent communication in a multi-agent reinforcement learning setting, where the agents solve cooperative tasks and have access to a communication channel. The communication channel may consist of either discrete symbols or continuous variables.
    We introduce an inductive bias to aid with the emergence of good %good? bedre ord? 
    communication protocols for continuous messages, and we look at the effect this type of inductive bias has for continuous and discrete messages in itself or when used in combination with reinforcement learning. 
    We demonstrate that this type of inductive bias has a beneficial effect on the communication protocols learnt in two toy environments, Negotiation and Sequence Guess.
\end{abstract}

\section{Introduction}
Although communication in multi-agent reinforcement learning (MARL) may emerge when the agents are given a communication channel that contains no predetermined communication protocol, efficiently doing so often turns out to be a difficult task. As is the case in MARL, the credit assignment problem together with the moving target problem leads to the existence of robust shadowed equilibria in many environments. [citations needed] As a result, when it comes to communication; instead of benefiting from learning a ``shared language'', the agents may end up disregarding the communication channel altogether and only act upon their own observation. 

In order to combat these issues and facilitate the emergence of meaningful communication, that is, communication which increases the expected return over a communication-free baseline, \citet{eccles2019biases_emergent} extended the measures of \textit{positive signaling} and \textit{positive listening} from \citet{lowe2019pitfalls} to be used as inductive biases.
The basic idea is to add additional terms to the loss function which encourage the following: In the case of positive signaling, a speaker is incentivized to produce different messages from different observations. While for positive listening a listener is incentivized to produce different actions from different messages. Overall, this ensures an improved use of the communication bandwidth and a reduced chance of not acting based upon messages. In the works by \citet{eccles2019biases_emergent}, this scheme is explored for when the messages passed between the agents are built from discrete symbols.

An orthogonal approach to improving the stability of learning to communicate is to allow for the gradient signal to flow through the communication channel. Doing so shifts the problem from a decentralized towards a centralized training paradigm, which helps at alleviating some of the issues plaguing the former\cite{multiagent_deep_RL_SURVEY}. The cost is that the gradient information needs to be available and passed between the agents while training, which may or may not be viable in a given setting. 

In this work, we look at how MARL agents learn to form a ``shared language'' in order to solve cooperative tasks. Using two toy examples, a variant of Negotiation \cite{toward_natural_turn-taking_in_a_Virtual_human_negotiation_agent, Emergent_Communication_through_Negotiation, deal_or_no_deal_end-to-end_learning_of_negotiation_dialogues, he2018decoupling_strategy_negotiation_dialogues}, and a new game we call Sequence Guess, we demonstrate how the positive signaling ideas can be extended to continuous communication protocols, where the discrete symbols are replaced with real numbers.           
We estimate the effect of continuous positive signaling on differentiable communication protocols \cite{CommNet, (BiCNet, autoencoderComm, zhu2022RLCOMM_survey}, reinforced communication protocols \cite{foerster2016DifferentiableBetterThanRIAL}, and a combination of both. The effect of a continuous communication protocol is also compared to the effect of a discrete one, in the otherwise discrete game of Sequence Guess.

\section{Positive Signaling}
We write the total loss function for a communicating MARL agent as 
$\mathcal{L} = \mathcal{L}_\text{rest} + \mathcal{L}_\text{comm}$, where the latter term is associated with communication and the former with other actions. The communication loss may be further subdivided into
\begin{equation}
    \mathcal{L}_\text{comm} = \mathcal{L}_\text{RC} + \lambda_{\text{IB}} \mathcal{L}_\text{IB}  
\end{equation}
where $\mathcal{L}_\text{RC}$ is the loss associated with the communication policy, $\mathcal{L}_\text{IB}$ is an inductive bias, and $\lambda_{\text{IB}} > 0$ is a weighting factor. We will focus on $\mathcal{L}_\text{comm}$ from here on.
Positive signaling is equivalent to maximizing the entropy of the average message policy, while at the same time minimizing the entropy of the message policy when conditioned upon a trajectory. We denote the policy for selecting message $m$ given a trajectory $x$ as $\pi(m|x)$, and the average message policy, weighted by frequency, as $\overline{\pi}(m) = \mathbb{E}_{x \sim \pi}[\pi(m|x)]$. $\overline{\pi}$ can be estimated from a mini-batch of $B$ trajectories by
\begin{equation}
    \overline{\pi}(m) \approx \frac{1}{B}\sum_b^B \pi(m|x_b),
\end{equation}
where the subscript $b$ labels the mini-batch members. Then, a natural inductive biaswhich encourages positive signaling would be
\begin{align}
    \mathcal{L}_\text{IB}(\pi, x) & = -\mathcal{H}(\overline{\pi}) + \lambda_\text{PS} \mathcal{H}(\pi(\cdot|x))  \\ 
    & = \sum_{m \in M} \overline{\pi}(m)\ln(\overline{\pi}(m)) -
     \lambda_\text{PS} \pi(m|x) \ln(\pi(m|x)) \nonumber, 
\end{align}
where $M$ is the set of all possible messages and $\lambda_\text{PS} > 0$ is a weighting factor.

In practice, however, minimizing the entropy when it is conditioned upon the current trajectory does not work well. One reason for this may be that for any $c < \log 2$ the space of policies with entropy at most $c$ is disconnected\cite{eccles2019biases_emergent}, in that the minimal possible entropy during a gradual message policy shift from an old to a new most likely message will at some point have to be at least
% message policy shift to a new most likely message would be 
$\log 2$, if the entropy of the initial message policy was less than $\log 2$. Because of this, Eccles et al. instead introduces a finite target entropy $H_\text{target} > 0$ for $\mathcal{H}(\pi(\cdot|x))$ and write
\begin{equation}
    \mathcal{L}_\text{IB}(\pi, x) = -\mathcal{H}(\overline{\pi}) + \lambda_\text{PS}(\mathcal{H}(\pi(\cdot|x)) - H_\text{target})^2
\label{eq: positive signaling}
\end{equation}
One can view $H_\text{target}$ as an exploration parameter, with a larger $H_\text{target}$ meaning a higher degree of exploration. 

\subsection{Positive Signaling for Continuous Communication}

We take a different approach to positive signaling in the case of the messages consisting of continuous variables rather than discrete symbols.
To make the computations viable, a compact support for the messages is needed.

\begin{figure}[htbp]
  \centering
  \includegraphics[width=0.75\linewidth]{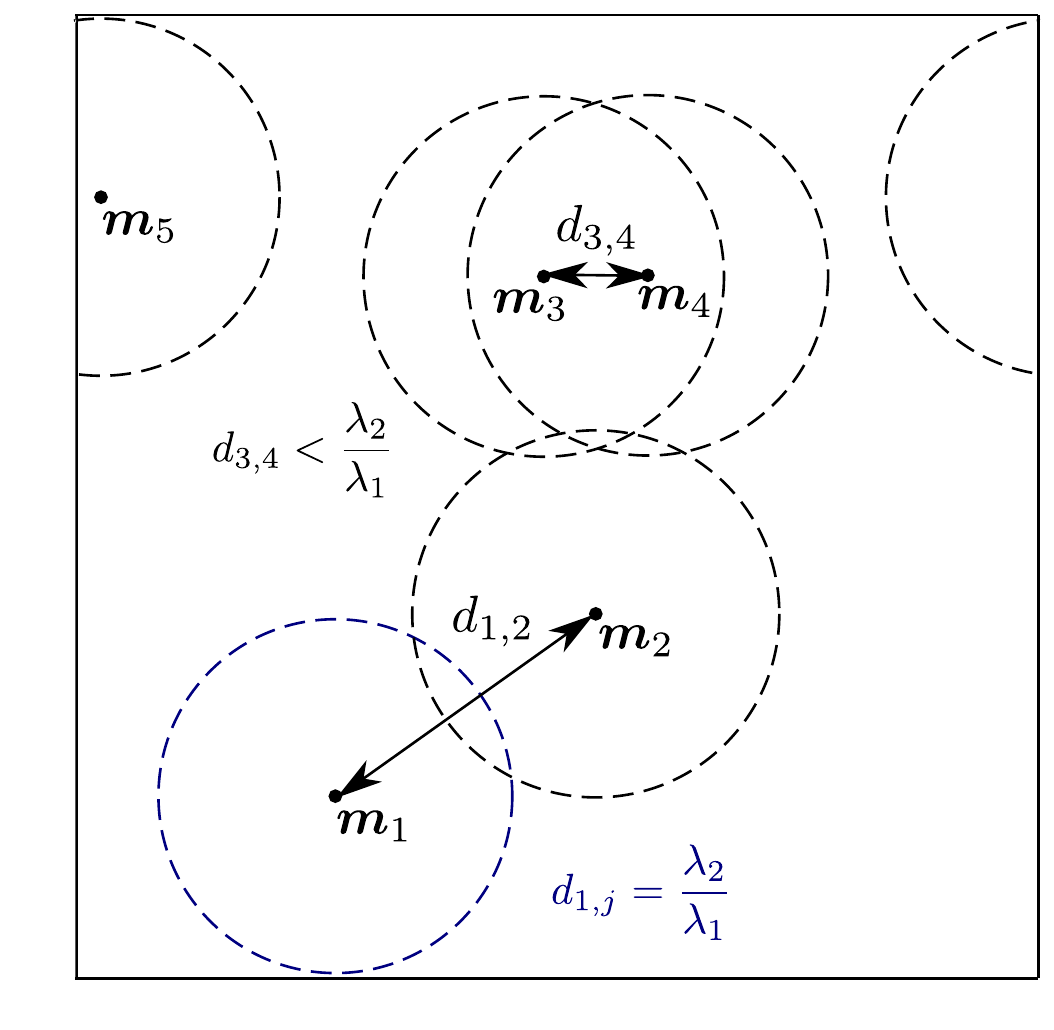}
  \caption{Example of continuous positive signaling where messages contain two components. The distance $d_{i,j} = d(\boldsymbol{m}_i, \boldsymbol{m}_j)$ is calculated between every message. The circles indicate where $d_{i,j} = \lambda_2/\lambda_1$, in other words where message $\boldsymbol{m}_i$ is considered too close to $\boldsymbol{m}_j$. $\boldsymbol{m}_3$ and $\boldsymbol{m}_4$ show a case where $d_{3,4} < \lambda_2/\lambda_1$. $\boldsymbol{m_5}$ illustrates that the space wraps around.} 
  \label{fig:ContinuousPositiveSignaling}
\end{figure}

A particularly easy set is the $n$-dimensional torus, where each component of a message lies on an interval of the real line with the end points identified, i.e. a circle. Furthermore, by interpreting the messages as points on a $n$-torus, we observe that a more uniform average policy may be encouraged by making the messages mutually ``repulsive''. In practice, this can be achieved stochastically by introducing a pair-wise repulsive ``potential'' $\ell$ between the members of a mini-batch of $B$ messages $\boldsymbol{M}$, leading to an inductive bias on the form
\begin{equation}
    \mathcal{L}_\text{IB}(\boldsymbol{M}) = \frac{1}{B^2} \sum_{\boldsymbol{m}, \boldsymbol{m'} \in \boldsymbol{M}} \ell(\boldsymbol{m}, \boldsymbol{m'}).
    \label{eq:continuous_bias}
\end{equation}
Since the messages are continuous, the gradient of the message policy with respect to its weights will be affected by this term as long as we choose an almost everywhere differentiable $\ell$. In this work, we have used the simple functional form
\begin{equation}
    \ell(\boldsymbol{m},\boldsymbol{m'}) = \max[-\lambda_1 d(\boldsymbol{m}, \boldsymbol{m'}) + \lambda_2, 0],
    \label{eq:ell}
\end{equation}
where $\lambda_1 > 0$ and $\lambda_2 > 0$ are hyperparameters and $d$ is a distance measure in message space. Assuming that $\boldsymbol{m}, \boldsymbol{m'} \in (-1,1)^n$ with $-1$ and $1$ identified as the same point, %the distance between two messages is given by
we define the distance between the two messages to be
\begin{equation}
    d(\boldsymbol{m}, \boldsymbol{m'}) = \sqrt{\sum_{i = 1}^n \min(|m_i - m'_i|, 2 - |m_i - m'_i|)^2},
\end{equation}
where $m_i$ is the $i$-th component of $\boldsymbol{m}$. See figure~\ref{fig:ContinuousPositiveSignaling} for an illustration. The intuition behind Eq.~\eqref{eq:ell} is to penalize two messages that are closer than $\lambda_2/\lambda_1$, and otherwise do nothing. Except for in pathological cases, this cutoff should not affect the inductive bias's push for a more uniform policy, while it does remove any constraining influence on the policy learning when messages are already far apart in message space. After all, what we are mainly interested in is to utilize the entire bandwidth of the message space, not to make the utilization perfectly uniform.

It should be noted that there are multiple ways of achieving a similar outcome as the one described here. Notably, an alternative to Eq.~\eqref{eq:continuous_bias} could read
\begin{equation}
    \mathcal{L}_\text{IB}(\boldsymbol{M}) = \frac{2}{B} \sum_{\substack{\boldsymbol{m} \in \boldsymbol{M}_A \\ \boldsymbol{m'} \in \boldsymbol{M}_B}} \ell(\boldsymbol{m}, \boldsymbol{m'}),
\end{equation}
where $\boldsymbol{M}$ has been equipartitioned into $\boldsymbol{M}_A$ and $\boldsymbol{M}_B$. This reduces the computational complexity of the inductive bias, at the expense of a higher variance. Furthermore, while the functional form of Eq.~\eqref{eq:ell} seems to work well in practice, it is straightforward to construct alternatives.

Since the effect of the latter term in Eq.~\eqref{eq: positive signaling} is to tune the degree of exploration for the message policy when conditioned upon a trajectory, we do not attempt to introduce such a term in the continuous case. Rather, we let the degree of exploration be controlled explicitly by the communication policy. 

\section{Experiments}
The main purpose of the experiments done here is to show how using continuous positive signaling for the communication protocol results in improved performance. We demonstrate this using two toy examples, a variant of \textit{Negotiation}, and \textit{Sequence Guess}, which are described below. For the experiments the REINFORCE algorithm \cite{ReinforceWilliams, Sutton} has been used, with a policy modeled as a neural network with a final softmax layer in case of discrete actions, and as a normal distribution with mean and standard deviation parameters given by a neural network when dealing with continuous actions. See appendices \cref{ap:Negotiation_Hyperparameter_Details,ap:Sequence_Guess_Hyperparameter_Details} for further details.

Pytorch has been used both to create the environments and the agents \cite{pytorch}. One experiment run on an NVIDIA GeForce RTX 3080 takes about one hour with roughly 40\% GPU utilization, this results in an estimated use of $3.2 \cdot 10^{16}$ FLOP during an experiment run.

\subsection{Negotiation}

\textit{Negotiation} is a game that consists of two agents, $A$ and $B$, which, through negotiation, try to distribute a set of $k$ different types of divisible items (e.g. beverages) among themselves. The game is fully cooperative and individual rewards are shared in order to produce a final reward for both agents.

\begin{figure}[htbp]
  \centering
  \includegraphics[width=\linewidth]{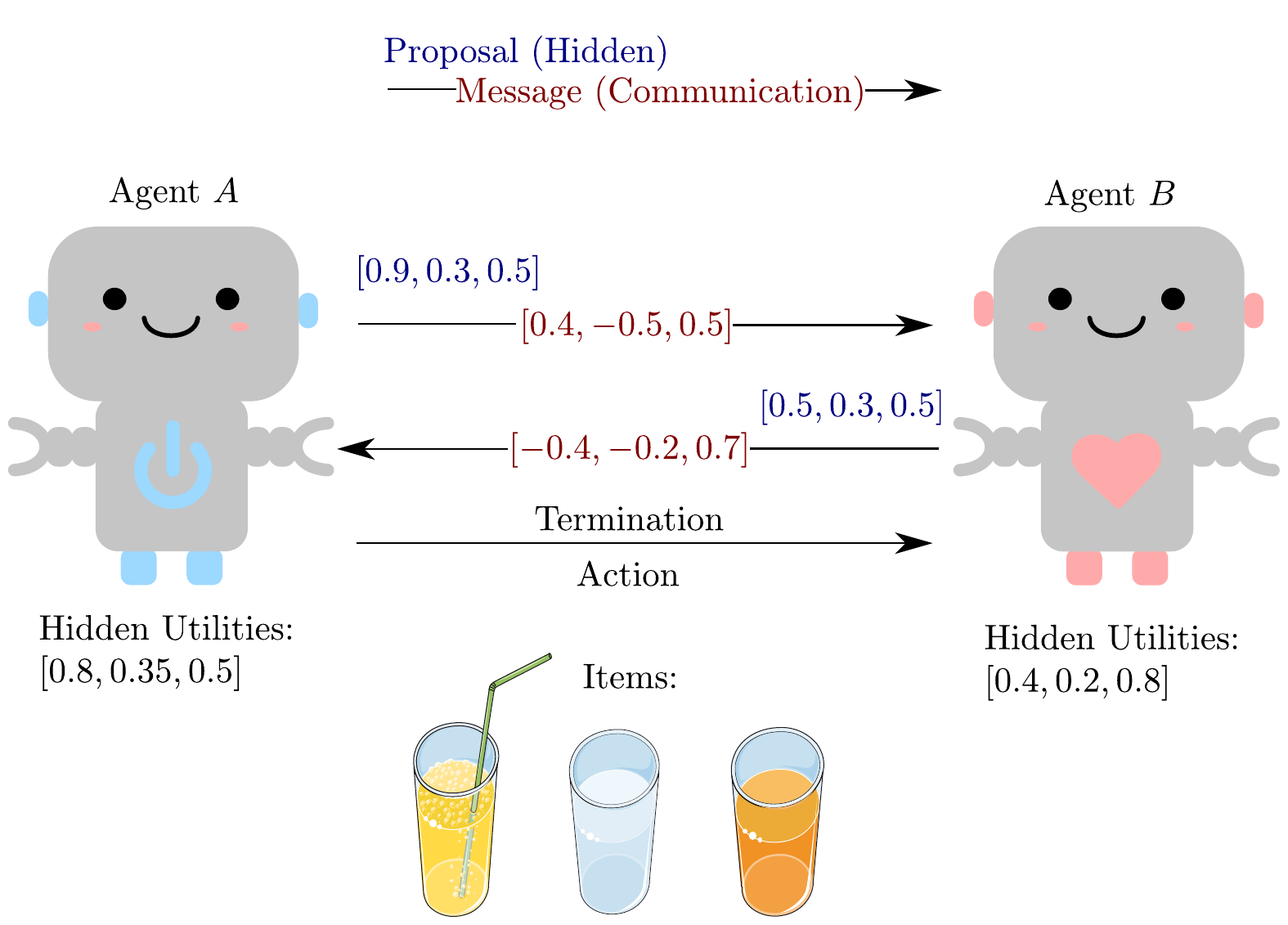}
  \caption{Negotiation. An example run of two agents negotiating over three different types of beverages. Messages have no predefined meaning. The hidden utilities indicate how each beverage is weighted when calculating the reward. A proposal of $[0.9, 0.3, 0.5]$ from agent $A$ would mean agent $A$ receives these proposed fractions of each beverage (here: soda, water and orange juice), while agent $B$ receives the remainder. Agent $B$ can either accept this proposal or come with a counter-proposal. In this example, where agent $A$ accepts $B$'s counter-proposal, the negotiation ends. The agents individual raw reward will be $0.5 \cdot 0.8 + 0.7 \cdot 0.35 + 0.6 \cdot 0.5 \approx 0.95$ and $0.5 \cdot 0.4 + 0.3 \cdot 0.2 + 0.4 \cdot 0.8 \approx 0.58$, which leads to a shared reward of $r_\text{Neg} \approx (0.95 + 0.58)/(0.8 + 0.35 + 0.8) \approx 0.78$.
  The robots have been taken from \cite{Robot1, Robot2}. The beverages have been taken from \cite{beverage_1, beverage_2, beverage_3}.}
  \label{fig:negotiationexample}
\end{figure}

At the beginning of a Negotiation round, the agents are individually assigned utilities $\boldsymbol{u} \in (0,1)^k$ for each of the items, drawn uniformly at random. They receive their own utilities as inputs, but are blind to the utility vector of their partner. The game proceeds in a turn-based manner by the agents alternately making a partitioning proposal $\boldsymbol{p} \in (0,1)^k$ and passing a message $\boldsymbol{m} \in (-1,1)^n$ to their partners, until either the partner agrees to the proposal, or a finite time limit $T$ has been reached. Note that the proposals are not explicitly shared, but have to be communicated through the messages, which in turn have to acquire meaning through collaborative learning. In addition to messages and utilities, the agents receive the current turn index $t$ as an input. A proposal agreement in the first turn, i.e. before the first agent has received a message, is ignored. The dynamics is illustrated in figure \ref{fig:negotiationexample} for $k = n = 3$.

At the end of a negotiation round, if the agents have come to an agreement, an individual normalized reward is calculated as 
\begin{equation}
     r_x = r(\boldsymbol{p}_x; \boldsymbol{u}_x) = \sum_{i=1}^k p_{xi} u_{xi},
\end{equation}
for $x \in \{A,B\}$, where $p_{Ai} + p_{Bi} = 1$. Note that if agent $A$ is the proposing agent, $\boldsymbol{p}_B$ is a function of the proposal $\boldsymbol{p}_A$, and not an independent proposal. The overall, shared reward is then defined as
\begin{equation}
    r_\text{Neg} = \frac{r_A + r_B}{r_\text{max}}.
\end{equation}
The scaling factor $r_\text{max} =
\sum_{i=1}^k \max(u_{Ai}, u_{Bi})$ is to ensure that $\max_{\boldsymbol{p}} r_\text{Neg} = 1$, regardless of the randomly drawn utilities. If the agents do not come to an agreement before reaching the time limit $T$, they are punished by $r_\text{Neg} = -1$. 

\subsection{Sequence Guess}
\label{sec:Sequence Guess}

We propose a new game, \textit{Sequence Guess}, inspired by the board game \textit{Mastermind}\cite{mathworld:mastermind}. Sequence Guess is a cooperative, asymmetrical two-player game consisting of a mastermind and a guesser. The goal is for the guesser to as quickly as possible guess a target sequence $\boldsymbol{a} = (a_1, a_2, \ldots, a_k)$ drawn from a finite alphabet $\Sigma$, while being guided by the mastermind through messages. Depending on the variant of the game, a message may consist of either a collection of discrete symbols or continuous variables. As with Negotiation, the meaning of a message is not predetermined, but has to be learned.

\begin{figure}[htbp]
  \centering
  \includegraphics[width=\linewidth]{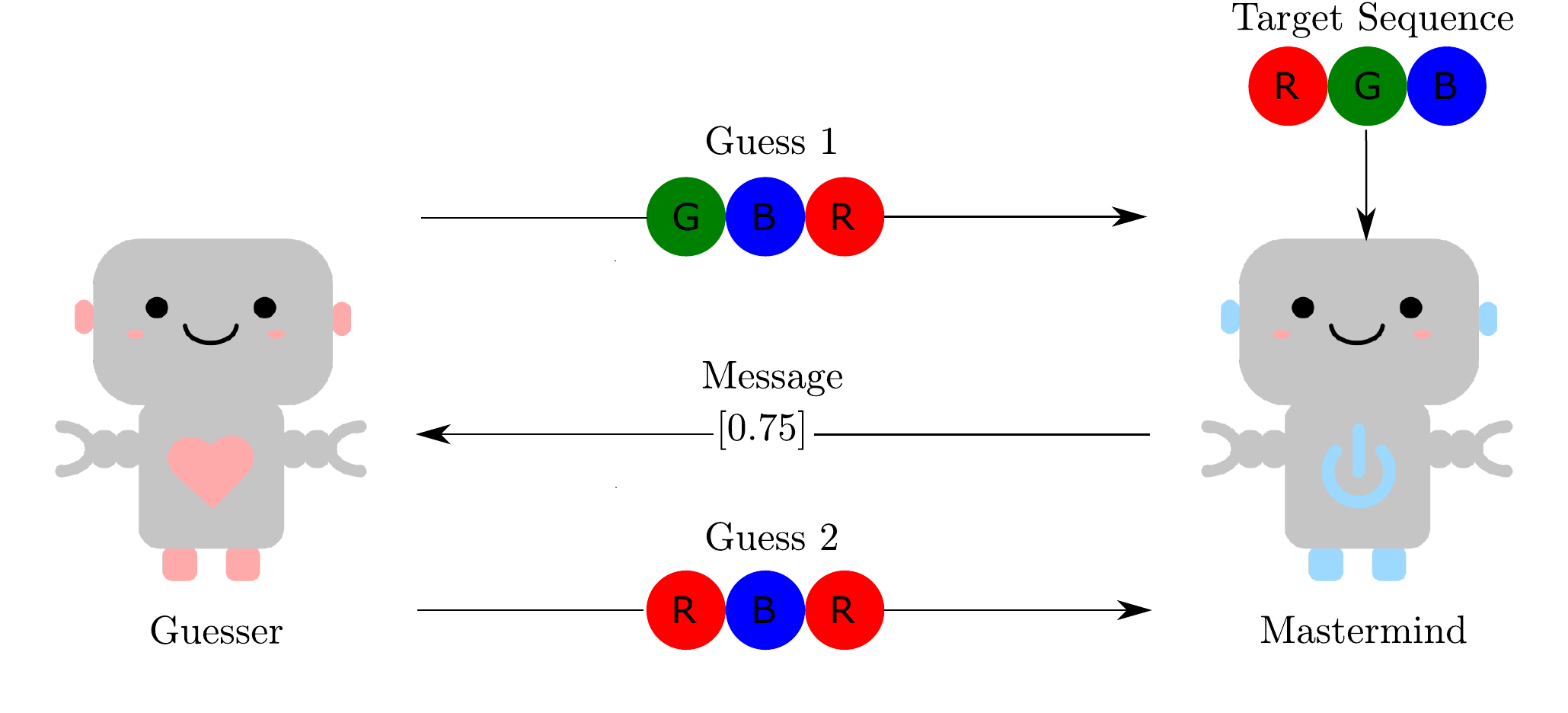}
  \caption{An excerpt of Sequence Guess. The guesser attempts to guess some target sequence, while the mastermind tries to provide information about the target sequence to the guesser. Here, the alphabet size is 3 and the target sequence length is 3, while messages consist of one real number. The robot figures are from \cite{Robot1, Robot2}.}
  \label{fig:seqguessexplained}
\end{figure}

Each turn of Sequence Guess consists of one sequence guess $\hat{\boldsymbol{a}} \in \Sigma^k$ by the guesser and one reply message from the mastermind: The guesser receives the latest message from the mastermind (initially a constant) as well as the current turn number $t$, and returns a guess. Then the mastermind replies with a new message based on the correct sequence, the guess, and the current turn number. If the guess is the same as the target sequence, or $t = T$, the game terminates.
Figure \ref{fig:seqguessexplained} shows an excerpt of a possible realization of Sequence Guess.

When a game of Sequence Guess ends, the agents are rewarded according to the fraction of letters in the guess sequence being equal to the corresponding letters in the target sequence, as well as a time penalty to encourage fast solutions:
\begin{equation}
    r_\text{Seq} =  -0.1t + \frac{1}{k} \sum_{i=1}^k \mathbf{1}_{a_i = \hat{a}_i}.
    \label{eq:sequence_guess_return}
\end{equation} 

Sequence Guess differs from Negotiation in a few key aspects. Negotiation is symmetric; each agent is given the same type of information, and the game is ``dialogue focused''. Sequence Guess, on the other hand, is ``monologue focused'', isolating the tasks of ``formulating'' and ``understanding'' messages to separate agents.
In Negotiation it is possible for the agents to learn a policy significantly better than the non-repeated random policy even with the communication channel removed. This is not possible in Sequence Guess.
Thus, Sequence Guess cannot suffer from shadowed equilibria.

\subsection{Results}
\begin{figure}
    \centering
    \begin{subfigure}[b]{0.49\linewidth}
      \centering
      \includegraphics[width=\linewidth]{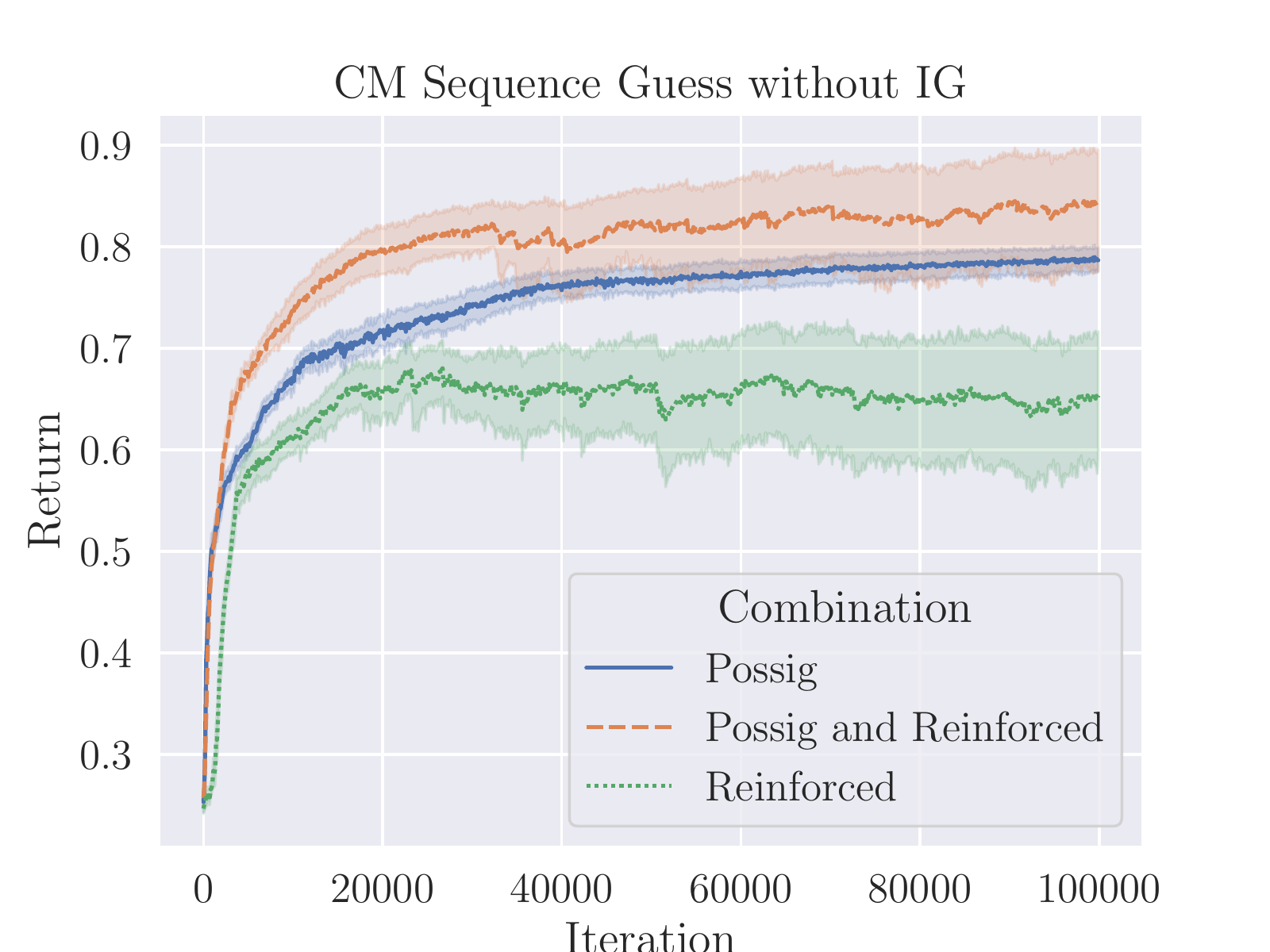}
      \caption{}
      \label{fig:RialConSeqGuess}
      \end{subfigure}
    %\hfill
    \begin{subfigure}[b]{0.49\linewidth}
      \centering
      \includegraphics[width=\linewidth]{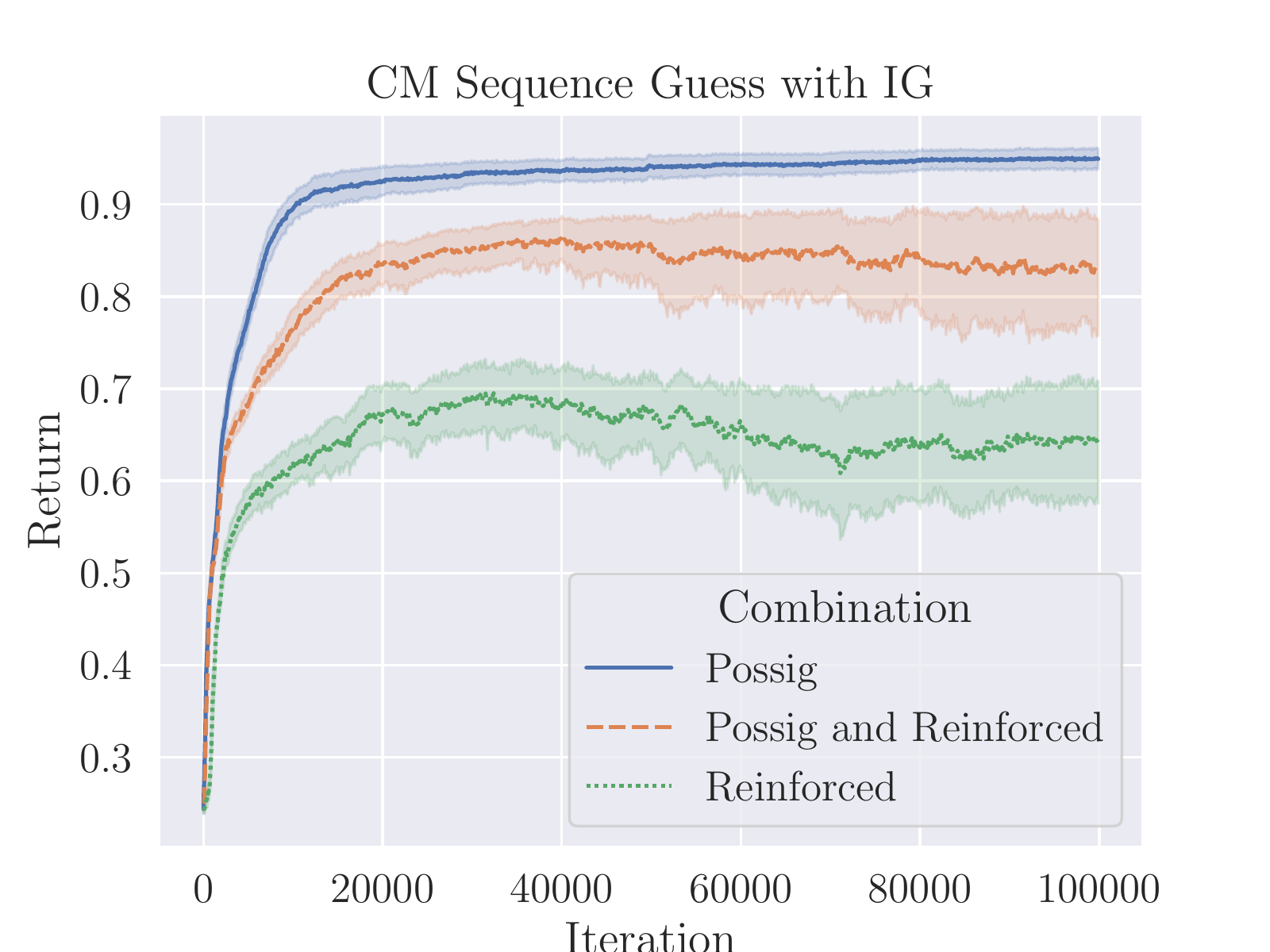}
      \caption{}
      \label{fig:DialConSeqGuess}
      \end{subfigure}
    \begin{subfigure}[b]{0.49\linewidth}
      \centering
      \includegraphics[width=\linewidth]{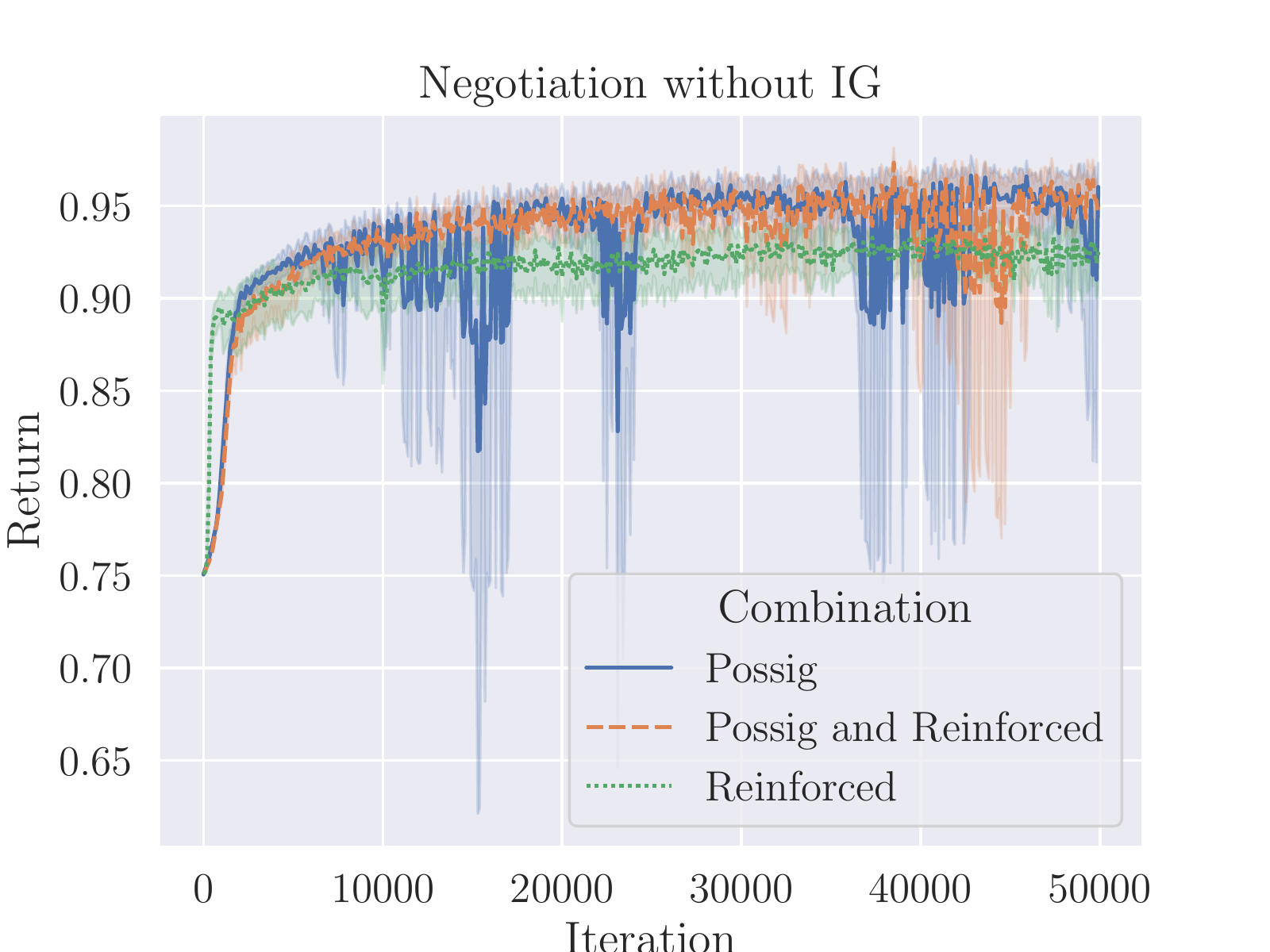}
      \caption{}
      \label{fig:RialNegotiation}
      \end{subfigure}
    \begin{subfigure}[b]{0.49\linewidth}
      \centering
      \includegraphics[width=\linewidth]{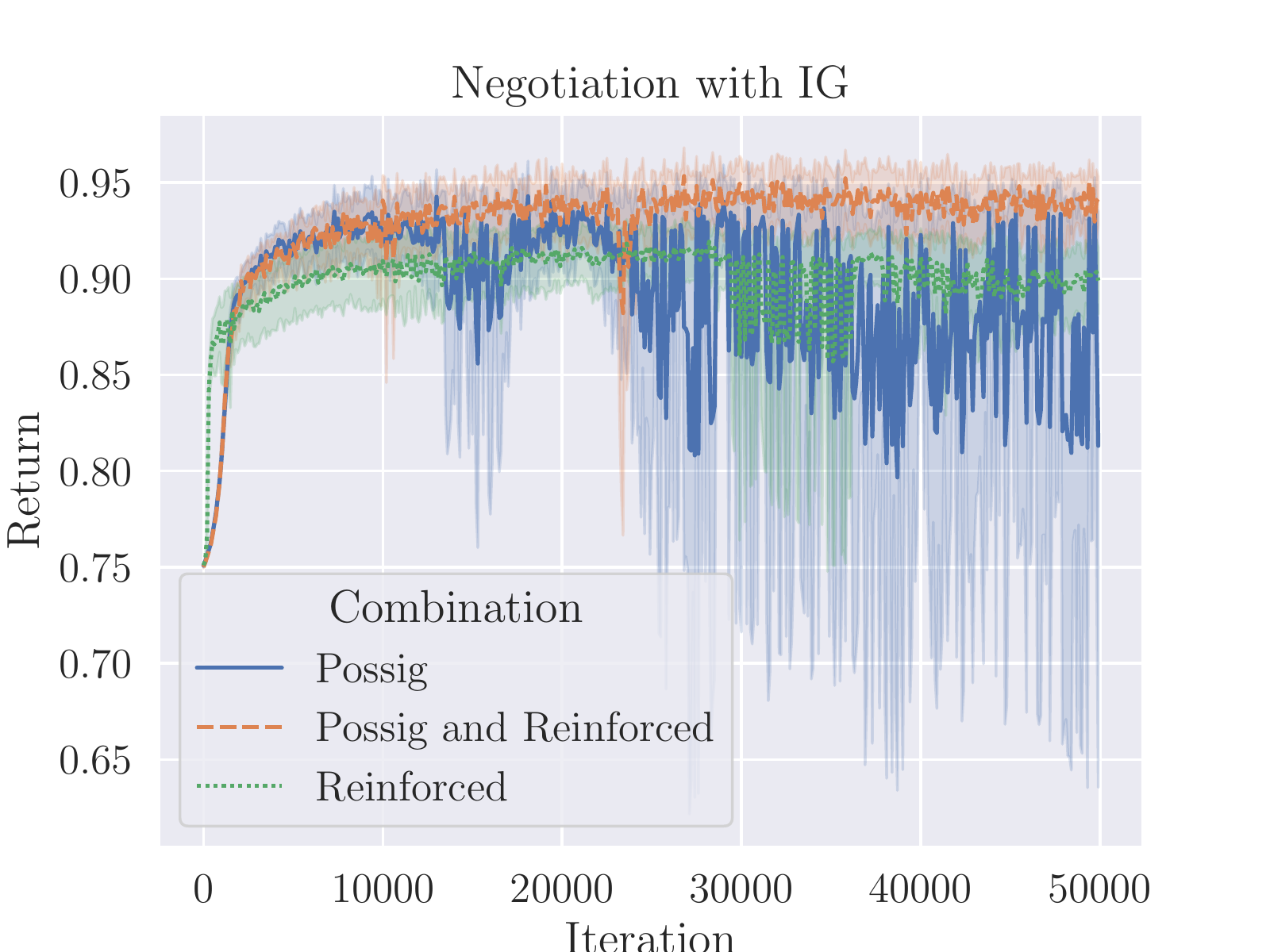}
      \caption{}
      \label{fig:DialNegotiation}
      \end{subfigure}
    \begin{subfigure}[b]{0.49\linewidth}
      \centering
      \includegraphics[width=\linewidth]{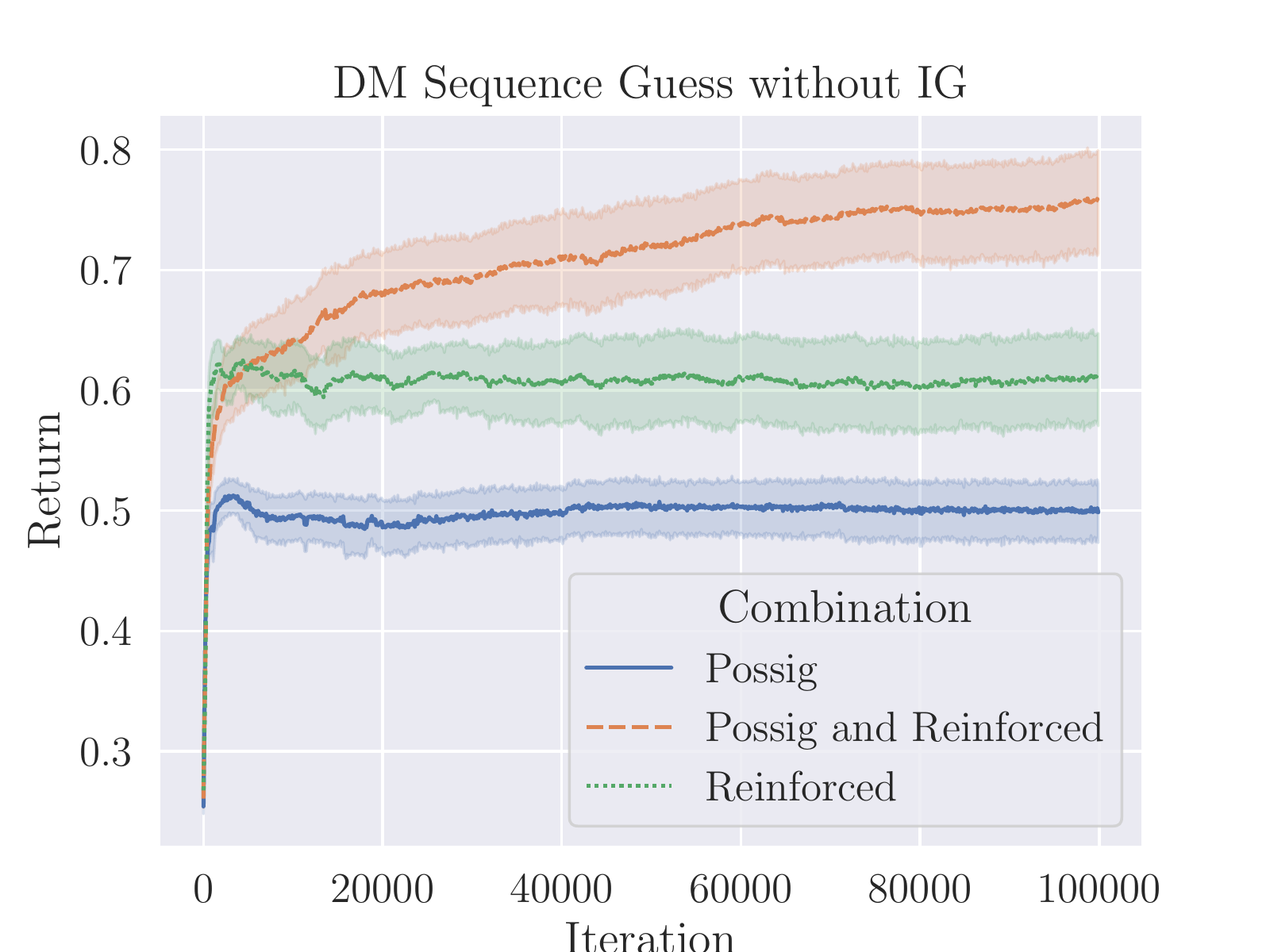}
      \caption{}
      \label{fig:RialDiscSeqGuess}
      \end{subfigure}
    \caption{
    Summary of 30 independent experiment runs for each game and loss function combination investigated. Each sample is an average of a mini-batch of size 2048. The lines indicate the means, while the bands denote $95\%$ confidence intervals.
    The rows display the different games, continuous message (CM) Sequence Guess, discrete message (DM) Sequence Guess, and Negotiation, with the columns are organized with respect to whether interagent gradients are allowed to flow or not. The returns have been scaled to ensure that their maximal expectation values are one. See the text for more details.
    }
    \label{fig:results}
\end{figure}

To demonstrate the effects of positive signaling, we set up a number of Negotiation and Sequence Guess ablation experiments, both when interagent gradients are allowed to flow and not. We focus on three cases: when the communication policy is learned using only a standard reinforcement learning approach without an inductive bias (RC), when it is obtained through minimizing the positive signaling inductive bias (PS), and when a combination of these two methods is used. The results of these experiments are shown in figure \ref{fig:results} and summarized in table \ref{table:results}. Note that the average returns of the Sequence Guess experiments have been shifted by a factor of $1-(1 \cdot 1/27+0.9\cdot 26/27)\approx 0.0963$ \footnote{The chance for the initial guess to be correct is $1/27$ which gives a reward of $1$, under an optimal policy the reward will be $0.9$ if the initial guess is incorrect.} to ensure that the expected return under an optimal policy is $1$. 

We observe that including a positive signaling learning bias typically improves the learning, as does interagent gradients. Moreover, continuous communication protocols outperform discrete ones. These points are not surprising, given that continuous messages and gradients carry more information compared to only communicating discrete messages. 

Based on these results, it may seem that positive signaling sometimes contributes more to learning a communication protocol than RC, or vice versa; see Figs. \ref{fig:RialConSeqGuess} and \ref{fig:DialConSeqGuess} versus Fig. \ref{fig:RialDiscSeqGuess}. One should, however, keep in mind that the positive signaling learning bias is different when using continuous and discrete messages.

Note that in \cref{fig:DialConSeqGuess} using only the positive signaling learning bias performs best. This may be due to the the fact that in this case all the information can be communicated in one message, making it possible for PS in of itself to push the learning towards an optimal bandwidth usage. In this case, adding an extra RC component will mostly provide detrimental noise. When the information cannot be communicated in a single message, this may not apply; then, a combination of PS and RC may perform better, as RC may help prioritize the most important information. Why this does not happen in \cref{fig:RialConSeqGuess} is not entirely clear to us, but we suspect that without a gradient to couple the learning of sending and receiving messages is less stable, hence in this case, an RC contribution may provide a helpful learning signal.

Without communication, the maximum possible expected return for Sequence Guess is $1/9 \approx 0.11$, while for Negotiation simulation experiments show that it is $\sim 0.92$. Assuming that it is easier to learn a non-communicative policy, this explains why the separation between the different Negotiation experiments (\cref{fig:DialNegotiation,fig:RialNegotiation}) is smaller than for the corresponding Sequence Guess experiments (\cref{fig:RialConSeqGuess,fig:DialConSeqGuess,fig:RialDiscSeqGuess}). 

The discrepancy between \cref{table:results} and \cref{fig:results} for Negotiation is due to averaged out oscillations; it seems that the learning is unstable with respect to whether the communication protocol is beneficial.

\begin{table}[]
\caption{
Summary of 30 independent experiment runs, each with a batch size of 2048. The table shows the average returns of the best mini-batches with $95\%$ confidence intervals, i.e. the expected highest return. \emph{IG}, \emph{CM}, and \emph{DM} refers to \emph{Interagent Gradients}, \emph{Continuous Messages}, and \emph{Discrete Messages}, respectively.}
\label{table:results}
\begin{tabular}{lllll}
\toprule
& & RC & PS & RC and PS \\ 
\midrule
\multicolumn{1}{l}{\multirow{3}{*}{IG}} & Negotiation &$0.942 \pm 0.009$   &$0.976 \pm 0.010$   & $0.979\pm 0.008$ \\
\multicolumn{1}{l}{}   & CM Sequence Guess & $0.780\pm 0.034$ & $0.958 \pm0.010$ & $0.921 \pm 0.020$                  \\ \midrule
\multicolumn{1}{l}{\multirow{3}{*}{No IG}} & Negotiation        & $0.951 \pm 0.008$ &$0.987 \pm 0.002$        & $0.986 \pm 0.003$                     \\
\multicolumn{1}{l}{}                                         & CM Sequence Guess     & $0.778\pm 0.029$          & $0.803\pm0.011$        &$0.897 \pm 0.024$                           \\
\multicolumn{1}{l}{}                                         & DM Sequence Guess & $0.693\pm0.020$            & $0.551\pm0.014$       & $0.782\pm0.035$                \\ \bottomrule
\end{tabular}
\end{table}

%%%%%%%%%%%%%%%
\section{Discussion and Summary}
%%%%%%%%%%%%%%%

In this work, we have shown that it is possible to generalize the idea of positive signaling as an inductive bias\cite{eccles2019biases_emergent} for learning communication with continuous messages. The advantages of introducing such an inductive bias to the loss function has been demonstrated on two toy MARL environments, Negotiation and Sequence Guess.

Positive signaling is encouraging the agents to learn to utilize the entire bandwidth of the communication protocol without getting stuck in local minima where two input states result in the same message. This is similar to what is done in variational autoencoders\cite{autoencoderComm}, where the average encoding is forced to follow a predefined distribution (typically a Gaussian), regularizing the learning space.

Continuous message protocols have some intrinsic advantages over discrete message protocols when it comes to learning: a smooth optimization landscape is usually preferable when dealing with policies modeled by neural networks. In addition, there is a practical limit on how large the communication alphabets can be, since their one-hot encoding grows linearly with the alphabet size. Furthermore, computing a positive signaling bias through ``repelling'' messages in a continuous message space is easier and more robust than estimating entropies based on mini-batches of discrete messages.

It is noteworthy that continuous message protocols seem to generally perform better than discrete message protocols, even when the desired information is discrete; compare \cref{fig:RialDiscSeqGuess} with \cref{fig:RialConSeqGuess,fig:DialConSeqGuess}.

If the set of possible states is smaller than the set of possible messages, loss-less communication is achievable and can be encouraged by using positive signaling. If the contrary holds, the sending agent can prioritize by adding RC. In practice, using only RC is likely not desirable, since the agents can get stuck in a local optimum where the same message is sent in two different states; most attempts at deviating from such a communication policy is likely to lead to a worse performance, driving the agents back to their original suboptimal policies. 

%%%%%%%%%%%%%%%
\section{Acknowledgements}
%%%%%%%%%%%%%%%
We would like to thank the Department of Informatics, University of Bergen, for financial support while completing this work.

\bibliographystyle{chicago}
\bibliography{possig}

\appendix
\section{Example PyTorch Implementation of Continuous Positive Signaling}

\begin{lstlisting}
    def positive_signalling_loss(self, means):
        means = means.repeat(len(means), 1, 1) 
        # Means has dimensions of length: 
        # [Batch Size, Batch Size, Sequence Length]
        means_2 = torch.transpose(means.clone().detach(), 0, 1)
        distances = torch.abs(means - means_2)
        delta = torch.minimum(distances, 2 - distances)
        delta = torch.pow(delta, 2))
        delta = torch.sum(delta, dim=2)
        delta = torch.triu(delta, diagonal=1)
        delta[delta == 0] = torch.inf
        delta = torch.sqrt(delta)
        zeroes = torch.zeros(delta.shape)
        loss = torch.maximum(-lambda_1 * delta + lambda_2, zeroes)
        loss = torch.mean(loss)
        return loss
\end{lstlisting}

\section{Network Architecture, Algorithm and Hyperparameters for Negotiation}
\label{ap:Negotiation_Hyperparameter_Details}
\if 0
        self.positive_signalling = False
        self.reinforce_commloss = True
        self.interagent_gradients = True  # One optimizer for both agents and gradients flow through the messages.

        self.n_jobs = 1  # Number of jobs for multiprocessing.
        self.n_runs = 30  # Number of times to repeat the run, n_runs should be divisible by n_jobs.
        self.batch_size = 2048
        self.num_iterations = 50_000  # number of iterations in a run
        self.lr = 0.001  # Initial learning rate
        self.lr_new = 0.0001  # Learning rate after threshold
        self.threshold = 0.9  # Threshold of AVG reward in batch to adjust LR
        self.weight_decay = 0.0001  # Weight decay for critic and agents
        self.max_turns = 6
        self.visible_proposals = False  # Agents receive each others proposal as input
        self.lambda_1 = 250  # Factor in Continuous positive Signalling loss
        self.lambda_2 = 10  # Factor in Continuous positive Signalling loss
        self.reward_sharing = 1  # Degree of reward sharing 1 - both receive the same reward, 0 no reward sharing
        self.clip_value = 1.0 # Clip value for gradient clipping

        # Number of items is currently 3
        # Number of floats in an utterance/message is currently 3
        # Number of agents is currently 2

        self.device = "cuda"  # which device to run the program on ie "cuda" or "cpu"
        self.root_folder = "NewNegotiationExperiment"
        self.current_folder = None
\fi
\subsection{Algorithm}
For Negotiation the REINFORCE algorithm is used with a parameterized baseline. The ADAM optimizer\cite{kingma2014adam} is used for both the baseline and the and policy. The baseline is centralized and receives the same input as the agent whose turn it is, learning was observed to be unstable without the use of a parameterized baseline.

\subsection{Network Architecture}
The network architecture for the agents and the baseline is the same unless otherwise stated:
\begin{enumerate}
    \item Input: Hidden utilities, current message, current turn/max turns
    \item LSTM layer with output size 100, Hidden state persists across one game of Negotiation.
    \item Fully Connected layer with output size 100 and Leaky ReLU activation
    \item Fully Connected layer with output size 1 and a tanh activation for the baseline in order to produce the return prediction, for the policy the output size is 13.
    \item For the policy 6 output variables are used as means ($\boldsymbol{\mu}$) 6 are used as standard deviations $\sigma(\boldsymbol{s})$, where $\sigma$ is the sigmoid activation function. $\boldsymbol{\mu}$ and $\sigma(\boldsymbol{s})$ are then used to initialize six normal distributions. Three normal distributions are used to sample the proposal $\sigma(\hat{\boldsymbol{y_1}})$. Three normal distributions are used to create the message $\tanh(\hat{\boldsymbol{y_2}})$. The remaining output is used to determine probability of termination $\sigma(\hat{y_3})$  from which the termination action is sampled. 
\end{enumerate}

\begin{table}
\centering
\begin{tabular}{@{}ll@{}}
\toprule
Hyperparameter    & Value   \\ 
\midrule
Batch Size        & 2048    \\
Iterations        & 50 000 \\
LSTM Layer Size      & 100     \\
Hidden Layer Size & 100 \\
Max Turns & 6 \\
$\lambda_1$ & 250 \\
$\lambda_2$ & 10 \\ 
Gradient Clip Value & 1 \\ 
$\beta_1$ (Decay Rate, Adam) & 0.9\\
$\beta_2$ (Decay Rate, Adam) & 0.999\\
Weight Decay & 0.0001 \\ 
\bottomrule
\end{tabular}
\caption{Hyperparameters used in Negotiation, the same layer sizes are used for the agents and the parameterized baseline}
\label{tab:neg-hyperparameters}
\end{table}

The learning rate $\alpha$ begins at 0.001 for both the baseline and the policies. When an iteration with $r \geq 0.9$ is reached, the learning rate is reduced to 0.0001 for both the baseline and the policies. $r \geq 0.9$ indicates a close to optimal joint policy if the agents only know their own hidden utilities. The reason for the learning rate adjustment is that the learning can very easily become unstable with a learning rate of 0.001. While learning would be very slow if it began with a learning rate of 0.0001. Table \ref{tab:neg-hyperparameters} shows the hyperparameters used. The hidden layer sizes, message sizes and number of item categories are the same ones that \citet{Emergent_Communication_through_Negotiation} used for their implementation. The choice of a ¨large batch" size is based upon \citet{attentionalHiddenStateSharing} who noted that a large batch size helped accelerate their learning process in a MARL setting.

\if 0
\subsubsection{Output of the Policy}
$\boldsymbol{p}$ is sampled from a normal distribution whose mean and standard deviation are outputs from the policy network, a sigmoid activation is used on the output for standard deviations in order to avoid negative numbers and large values for the standard deviation. After sampling from the normal distribution the sigmoid activation function is used in order to ensure a legal proposal.
$\boldsymbol{m}$ is sampled in the same manner as $p$ except that the $\tanh$ activation function is used to define the boundaries of a message, if the communication channel isn't bounded then numerical stability will be an issue as the policy may increase or decrease the value of a message indefinitely.
\fi

The network also outputs the probability of termination from a sigmoid activation, and the termination action is sampled from this probability.

\if 0
\subsubsection{Details on the Baseline}
Since learning has been observed to be unstable we have decided to use a parameterized baseline that is potentially able to accurately predict the expected reward from a given state, in order to reduce variance in the reward signal.

The baseline receives the current turn, current proposal and current message as input. The baseline estimates the reward for the player whose turn it is, this is done because degree of reward sharing is hyper parameter that is not included here.
 When calculating the loss for the parameterized baseline Mean Squared Error is used.

The inclusion of a centralized baseline does not affect the idea of the agent's only being allowed to communicate through a communication protocol, because during an episode no information is shared between the agents through the baseline.
\fi

\section{Network Architecture, Algorithm and Hyperparameters for Sequence Guess}
\label{ap:Sequence_Guess_Hyperparameter_Details}
\if 0
        self.positive_signalling = True
        self.reinforce_commloss = False
        self.interagent_gradients = False  # One optimizer for both agents and gradients flow through the messages.
        self.continuous = True
        self.lr = 0.0001

        self.n_jobs = 1  # Number of jobs for multiprocessing.
        self.n_runs = 30  # Number of times to repeat the run, n_runs should be divisible by n_jobs.
        self.batch_size = 2048
        self.num_iterations = 100_000  # number of iterations in a run
        self.guess_alphabet_size = 3  # Alphabet size of the target language
        self.guess_seq_length = 3  # Length of the target
        self.utt_alphabet_size = 3  # Alphabet size for message language, applies to discrete sequence guess
        self.utt_seq_length = 3  # Length of the message
        self.max_turns = 3
        self.basline_constant = 0.7  # Baseline for reward baseline = 0.7*baseline+0.3*return
        self.target_entropy_constant = 0.1  # 1 gives the maximum possible entropy 0 is no entropy
        self.lambda_1 = 100  # Factor in Continuous positive Signalling loss
        self.lambda_2 = 10  # Factor in Continuous positive Signalling loss
\fi
\subsection{Algorithm}
The REINFORCE algorithm with a baseline and the ADAM\cite{kingma2014adam} optimizer is used. A moving average is used as the baseline. This is how the baseline is calculated at iteration $t+1$:
\begin{equation}
    b_{t+1} = 0.7b_t + 0.3G_t
\end{equation}
Where $G_t$ is the mean return over the entire mini-batch for the current iteration. This is the same baseline that \citet{Emergent_Communication_through_Negotiation} used. Since the stability issues observed in \emph{Continuous Negotiation} has not been observed in this game, a more simple baseline should suffice.

\subsection{Network Architecture}
\label{Sequence Guess Network Architecture}
\subsubsection{Discrete messages}
\begin{figure}[h]
  \centering
  \includegraphics[width=\linewidth]{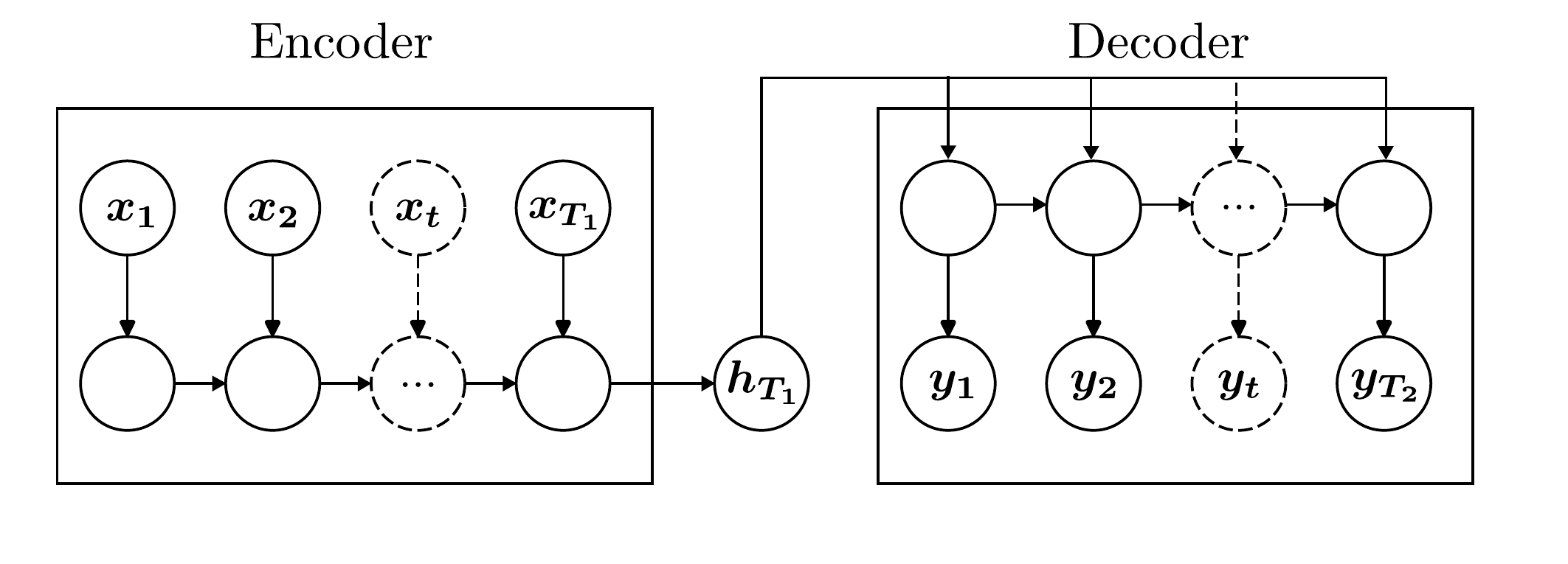}
  \caption{The encoder-decoder architecture used for DM Sequence Guess. \emph{In the case of the mastermind} input $x_t$ will contain symbol number $t$ from the guess sequence and target sequence, $T_1$ will be the length of the target sequence and $T_2$ will be the length of the message sequence. The output $y_t$ is used in a fully connected layer with a Softmax activation function in order to find message symbol number $t$. 
  \emph{In the case of the guesser} input $x_t$ will contain symbol number $t$ from the message, $T_1$ will be the length of the message sequence and $T_2$ will be the length of the target sequence. $y_t$ is used in a fully connected layer with a Softmax activation function in order to find guess symbol number $t$. In both cases a one-hot encoding of the current turn is appended to the final hidden state of the Encoder in order to produce the context vector.} 
  \label{fig:seqguessarchitecture}
\end{figure}

Both the guesser and mastermind:
\begin{enumerate}
    \item Input Guesser: Current Turn, Last Message. Input Mastermind: Current turn, Last Guess, Target Sequence.
    \item An encoder-decoder using LSTMs, see \cref{fig:seqguessarchitecture} for details. Outputs a tensor of dimensions [Batch Size, Sequence Length, Alphabet Size]
    \item Fully Connected layer of input size 100 and output size 3 with softmax activation over final dimension.
    \item Initialize Batch Size $\times$ Sequence Length categorical distributions with "Alphabet Size" categories and sample the Guesses or Messages for the Guesser and Mastermind respectively.
\end{enumerate}
In the discrete case both agents are parameterized by an encoder-decoder architecture using LSTMs. A one hot encoding of the current turn is appended to the hidden state.  Figure \ref{fig:seqguessarchitecture} illustrates the encoder-decoder architecture in more detail. The decoder output is used in a fully connected layer of size 100. Guesses and messages are sampled according to the distribution from the softmax activation function in the final output layer.
\subsubsection{Continuous messages}
For continuous messages the architecture is the same except for that mastermind employs only the encoder part, where the final hidden state is used in a fully connected layer of size 100, then a ReLU activation and another fully connected layer, the output from this layer is used in the same manner as in Negotiation in order to generate the message.
The guesser employs only the decoder part for continuous messages, as messages are of the same type as those used for Negotiation.

\begin{table}
\centering
\begin{tabular}{@{}ll@{}}
\toprule
Hyperparameter    & Value   \\ \midrule
Batch Size        & 2048    \\
Learning Rate     & 0.001   \\
Iterations        & 100 000 \\
Encoder Size      & 100     \\
Decoder Size      & 100 + $n$ turns     \\
Hidden Layer Size & 100   \\
$\lambda_1$ & 100 \\
$\lambda_2$ & 10 \\ 
$\lambda_\text{PS}$ & 1 \\
$\lambda_\text{IB}$ & 1 \\
$\beta_1$ (Decay Rate, Adam) & 0.9 \\
$\beta_2$ (Decay Rate, Adam) & 0.999 \\
weight decay & 0.0001 if CM 0 if DM \\
\bottomrule
\end{tabular}
\caption{Hyperparameters used in Sequence Guess. The same hyperparameters are used for the guesser and the mastermind. The reasoning behind the choice of hyperparameters is they are of the same magnitude as the ones used by \citet{Emergent_Communication_through_Negotiation}. Note that weight decay with DM fails to produce any convergence, the most likely reason beings that the gradient signal in the initial correlations giving rise to communication is too weak compared to the gradient signal of weight decay.}
\label{tab:seq-guess-hyperparameters}
\end{table}

Table \ref{tab:seq-guess-hyperparameters} shows the hyperparameters used in Sequence Guess.

\end{document}